\documentclass[11pt]{article}
\usepackage{coling2018}
\usepackage{times}
\usepackage{url}
\usepackage{latexsym}

\usepackage{graphicx}
\usepackage{alltt}
\usepackage{array}

\title{Multi-task dialog act and sentiment recognition on Mastodon}

\author{Christophe Cerisara, Somayeh Jafaritazehjani, Adedayo Oluokun and Hoa T. Le\\
  Universit\'e de Lorraine, CNRS, LORIA, F-54000 Nancy, France \\
  {\tt cerisara@loria.fr, hoa.le@loria.fr} 
  \\}

\date{}

\begin{document}
\maketitle
\begin{abstract}
Because of license restrictions, it often becomes impossible to strictly reproduce most research results 
on Twitter data already a few months after the creation of the corpus. This situation worsened gradually as time passes and
tweets become inaccessible. This is a critical issue for reproducible and accountable research on social media.
We partly solve this challenge by annotating a new Twitter-like corpus from an alternative large
social medium with licenses that are compatible with reproducible experiments: Mastodon.
We manually annotate  both dialogues and sentiments on this corpus, and train a multi-task hierarchical recurrent
network on joint sentiment and dialog act recognition.
We experimentally demonstrate that transfer learning may be efficiently achieved between both tasks,
and further analyze some specific correlations between sentiments and dialogues on social media.
Both the annotated corpus and deep network are released with an open-source license.
\end{abstract}

\section{Introduction}

%
% The following footnote without marker is needed for the camera-ready
% version of the paper.
% Comment out the instructions (first text) and uncomment the 8 lines
% under "final paper" for your variant of English.
% 
\blfootnote{
    %
    % for review submission
    %
    % \hspace{-0.65cm}  % space normally used by the marker
    % Place licence statement here for the camera-ready version. See
    % Section~\ref{licence} of the instructions for preparing a
    % manuscript.
    %
    % % final paper: en-uk version 
    %
    % \hspace{-0.65cm}  % space normally used by the marker
    % This work is licenced under a Creative Commons 
    % Attribution 4.0 International Licence.
    % Licence details:
    % \url{http://creativecommons.org/licenses/by/4.0/}
    % 
    % final paper: en-us version 
   
    \hspace{-0.65cm}  % space normally used by the marker
    This work is licensed under a Creative Commons 
    Attribution 4.0 International License.
    License details:
    \url{http://creativecommons.org/licenses/by/4.0/}
}

Social media are a gold mine for researchers in many domains and especially in natural language processing,
because of the endless stream of linguistic content produced every day.
However, a major issue faced by every researcher working with Twitter data concerns the accessibility of datasets
previously extracted. Indeed, license restrictions limit the possibility to store tweets in a database for a long
period of time, and the proportion of tweets that are continuously deleted from the Twitter company servers 
makes any Twitter corpus quickly obsolete and impossible to retrieve after a few months.
This is a very serious issue for the ethics of experimental research, of which reproducibility is a foundational
feature.
Despite these limitations, many research and development works use Twitter data for
a wide variety of applications.

We propose in this work to exploit another social medium with language data, the Mastodon network,
which closely resembles Twitter, except for two important differences: Mastodon is a decentralized social network,
and it adopts permissive licenses that are compatible with reproducible research.
In particular, everyone may create his own Mastodon server with his cohort of registered users, and have his server
automatically federated within the worldwide social network.
To enable this, the Mastodon software is released with the open-source AGPL license, and the user-generated content
in the Mastodon network typically follows a Creative Commons license\footnote{see, e.g., {\url{https://forum.etalab.gouv.fr/tos#3}}}, which allows redistribution of posts.

Mastodon is very similar to Twitter with regard to content and usage, except that the user posts are limited by default to 500
characters, which gives the user the possibility to write longer sentences than on Twitter.
From a sociological point of view, Mastodon users are mostly composed of people who are either attracted by the technical challenge of setting up their own server,
or who have been disappointed by Twitter, sometimes because of advertisements, changes of policies and privacy threats,
or who feel a strong sense of belonging to a minority group, such as LGBT, because Mastodon naturally enables and favors the emergence of local communities.
Another difference, which is relevant for researchers in Linguistics and Natural Language Processing, is that
Mastodon is more international than Twitter, in the sense that the majority of Mastodon servers are
located in Japan and in the West of Europe (France, Spain and Germany in particular), and although
these servers are federated together, each of them brings together a community, which is often linguistic.
It is thus frequent to observe on some server a majority of posts that are not written in English.

In terms of the amount of language data produced every day, 
Mastodon is still less developed than
Twitter, but it is growing and is already large enough (with one and a half million of users as of February 2018)
to be useful for most research works.
We demonstrate this by scrapping and annotating a corpus of dialogues in English from Mastodon
and training and validating a deep learning model on this data.
Both the corpus and software described in this work are distributed freely
at {\url{https://github.com/cerisara/DialogSentimentMastodon}}.
We focus in this work on two aspects of Mastodon posts: dialog acts and sentiment recognition,
and on the correlations between both of these tasks.

One of the most studied natural language processing task on Twitter is sentiment recognition,
which is for example a recurrent task every year at the SEMEVAL evaluation campaign~\cite{semeval16,semeval}.
We thus have also annotated our Mastodon corpus with sentiments, in order to maximize
the potential impact and usage of this new corpus.
Conversely, there are fewer works that study dialogues on Twitter.
Dialogues on social media have quite a different form than on other media, such as vocal, sms
and meetings interactions, because the participants in the dialogue are often unknown in advance, 
they may come and leave at any time and live in different time zones.
The structure of social media dialogues thus typically forms a tree, or a directed acyclic graph if
we take into account @-mentions.
We focus next on dialog acts, also known as speech acts, which characterize the function of a phrase in the
course of a dialog. Typical dialog acts are questions, answers, disagreements...

The scientific hypothesis that is studied in this work is that sentiments are correlated to dialog acts
on social media, and that this correlation may be exploited to enable transfer learning between both tasks.
Intuitively, consider a dialog where user A writes something {\it positive} (sentiment) about a given smart phone brand.
Then, user B {\it disagrees} (dialog act) and points out a {\it negative} aspect of this brand.
An obvious correlation between dialog acts and sentiments can be found on this trivial example.
Surprisingly, the majority of works about sentiment analysis ignores such relations
between dialog acts and sentiments.
We propose next to demonstrate experimentally that this correlation is strong enough to enable
transfer learning between both tasks, and we further analyze 
both quantitatively and qualitatively some of the observed correlation patterns.

\section{Corpus annotation}

We hereafter call {\it post} the Mastodon equivalent of a tweet, i.e.,
a single message that is limited to 500 chars maximum.
From the user perspective, Mastodon is very similar to Twitter, and we can thus find
the same type of content that is found on Twitter.
We have crawled about 800,000 posts from the {\it octodon.social} Mastodon instance,
which is one amongst thousands of existing instances,
and have filtered out non English posts automatically with the {\it langdetect} python library.
We have then followed the {\it reply-to} links to structure all posts into dialog trees:
dialogues in social media are structured as trees, because anyone can get involved in a dialog
from any post that composes the dialog so far.
About half of the dialogues have two posts, $1/4$ three, and so on.
The longest dialog is composed of 44 posts.
The tree-dialogues are then split into a training and test set.

Before annotation, all dialogues are linearized, following the work of Zarisheva and Scheffler~\shortcite{scheffler}, which means that each branch of
the tree forms a unique dialog.
Two students with a Master degree in linguistics and fluent in English independently assigned two tags to each post in a dialog:
\begin{itemize}
    \item A sentiment tag with 3 possible values: positive (26\% of the corpus), negative (31\%) and neutral (43\%); The baseline performances, when always classifying posts as neutral, are F1=24.4\%.
    \item A dialog act with 27 possible values (the bold labels on the right of Table~\ref{tab:das}).
\end{itemize}
A typical dialog is shown in table~\ref{tab:ex}.

\begin{table}[h!t]
    \begin{center}
{\small{
\begin{tabular}{|c|c|c|m{10cm}|}
\hline
& {\bf Sent} & {\bf DA} & {\bf Textual content of segment}\\
\hline
\hline
    0& -&I& because this is getting way too much attention it wasn't even that funny URL\\
\hline
    1& +&I& LMAO \\
\hline
    2& *&O& why you still on twitter though ? \\
\hline
    3& -&W& i'm trying to get all my other friends to get on here before i deactivate ! \\
\hline
    4& -&I& I might have to do that too . Some ain't migrated yet . It's an issue . \\
\hline
    5& -&I& some of my mutuals are waiting for more leftists go get on here \\
\hline
    6& *&O& and i'm like 🙄  what are you waiting for ? ? ? \\
\hline
    7& -&I& then again i did have to spend hours to figure out how to work this site haha \\
\hline
    8& -&Q& lol was it that hard ? ? \\
\hline
    9& -&W& i'm not that good with technology or websites \\
\hline
    10& -&A& so yeah .. \\
\hline
    11& -&A& well yeah , the whole instances thing is kind of confusing tbh . \\
\hline
\end{tabular}
}}
\caption[]{Example of a typical dialog from the Mastodon corpus.}
\label{tab:ex}
\end{center}
\end{table}

The dialog act tags have been derived from the seminal work on the Switchboard corpus~\cite{DAMSL}
and have been adapted to take into account specificities of social media, taking inspiration from the work of Zarisheva and Scheffler~\shortcite{scheffler}.
The recently proposed ISO standard~\cite{iso} has also been taken into account in the design phase of the annotation guide.
This process has lead to the definition of the tags listed in Table~\ref{tab:das}.
After a first round of annotations and in order to avoid rare labels, these 27 tags have been further merged into 15 tags, whose distribution is shown in Table~\ref{tab:das}.

\begin{table}[h!]
{\footnotesize{
\begin{alltt}
- Communicative functions
  - General purpose functions
    - Information seeking functions
      - Question
        - Propositional question
          - Yes/No question \dotfill {\bf Q} (8.3\%)
          - Check question \dotfill (merged a priori in Q)
        - Set question
          - Wh* / Open question \dotfill {\bf O} (7.4\%)
        - Choice question
          - Open with choices \dotfill (merged a priori in O)
    - Information providing functions
      - Inform
        - Statement \dotfill {\bf I} (49.3\%)
        - Agreement \dotfill {\bf A} (7.9\%)
        - Disagreement \dotfill {\bf D} (1.9\%)
          - Correction \dotfill (merged a priori in D)
        - Answer
          - Open + choice answer \dotfill {\bf W} (9.9\%)
          - Confirm answer \dotfill {\bf Y} (merged in A)
          - Disconfirm answer \dotfill {\bf N} (merged in D)
    - Commissive functions
      - Offer \dotfill {\bf E} (1.4\%)
        - Promise \dotfill (merged a priori in E)
    - Directive functions
      - Request \dotfill {\bf R} (3.3\%)
        - Instruct \dotfill (merged a priori in R)
      - Suggest \dotfill {\bf S} (3.0\%)
    - Directive & commissive functions
      - Accept offer request suggest \dotfill {\bf P} (merged in A)
      - Decline offer request suggest \dotfill {\bf L} (merged in D)
  - Feedback functions
    - auto-positive acknowledgement \dotfill {\bf F} (0.2\%)
    - allo-positive acknowledgement \dotfill (merged a priori in F)
    - auto-negative acknowledgement \dotfill {\bf B} (merged in F)
    - allo-negative acknowledgement \dotfill (merged a priori in B)
  - Social obligation functions
    - Initial greetings \dotfill {\bf H} (2.0\%)
    - Return greetings \dotfill (merged a priori in H)
    - Initial goodbye \dotfill {\bf G} (merged in H)
    - Return goodbye \dotfill (merged a priori in G)
    - Apology \dotfill {\bf X} (merged in M)
    - Accept apology \dotfill {\bf C} (merged in A)
    - Thanking \dotfill {\bf T} (2.0\%)
    - Accept thanking \dotfill {\bf K} (merged in A)
- Exclamation \dotfill {\bf J} (1.5\%)
- Explicit performative ({\it hope you get better}) \dotfill {\bf V} (1.6\%)
- Sympathy ({\it I'm sorry to hear that}) \dotfill {\bf M} (0.6\%)
- Miscellaneous / other \dotfill {\bf *} (absent)
- UNK (too ambiguous to decide) \dotfill {\bf U} (removed)
- Malformed input \dotfill {\bf Z} (removed)
\end{alltt}
}}
\caption[]{List of dialog acts with the 27 labels used by annotators (letters in bold) and their distribution in the corpus (percentage within parentheses). Some labels have been merged a priori (before the annotation process), in order to simplify the annotation process. Very rare labels have been merged after the annotation process. The remaining 15 labels used to compute the F1 metric are the ones with a distribution percentage. The baseline recognition performance obtained when always answering {\bf I} is F1=35\%.}
\label{tab:das}
\end{table}

The inter-annotator agreement is 88.6\% for dialog acts and 90.2\% for sentiments.
The Cohen's kappa coefficient is 85.1\% for dialog acts and 90.2\% for sentiments.

The final training corpus is composed of 239 dialogues for a total of 1075 posts, and the test corpus of
266 dialogues for a total of 1142 posts. The vocabulary size is 5330 words.

\section{Multi-task model}

\subsection{Model description}

\begin{figure*}
\begin{center}
\includegraphics[width=12cm]{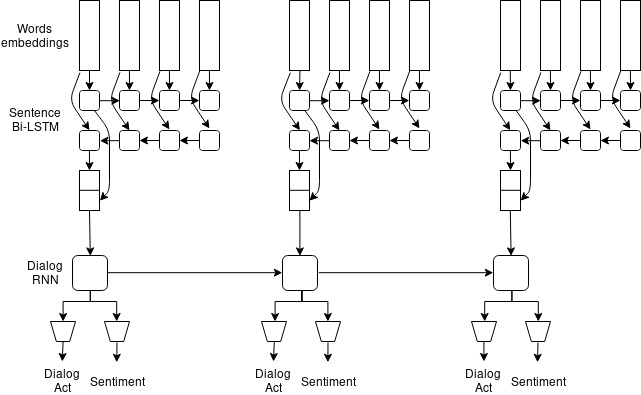}
\end{center}
\caption[]{Multi-task hierarchical recurrent model. The number of posts per dialog and the number of words per post may vary and are handled dynamically without any padding.}\label{fig:model}
\end{figure*}

The proposed model is shown in Figure~\ref{fig:model}.
It is a two-level hierarchical recurrent network similar to the one proposed in Sordoni et al.~\shortcite{sordoni15}:
\begin{itemize}
\item The first level ({\it post level}) takes as input the sequence of word embeddings in a post.
This first level is composed of a bi-LSTM, which outputs a single vector per post.
\item The second level ({\it dialog level}) takes as input the sequence of vectors produced at the first level, i.e.,
one vector per post for every post within a dialog.
The second level is composed of a standard RNN, because the length of dialogues rarely exceeds 10 posts and
so the LSTM cells do not offer a clear advantage over standard recurrent cells.
The output of this RNN at every post is passed to two MLPs, one for dialog acts and another for sentiment labels.
\end{itemize}

\subsection{Training procedure}

Two cross-entropy losses are used to train the model, one for dialog act recognition
and another for sentiment classification.
In our corpus, every post is manually annotated with both a sentiment and dialog act labels.
So standard multi-task training simply consists in backpropagating the gradient of both losses with equal weights.
However, we also realize some transfer learning experiments with fewer annotations in one of the tasks.
In such cases, we artificially remove the gold sentiment label for training and only
backpropagate the gradient of the dialog act loss, and vice versa when transferring from the sentiment to the dialog act task.

A development corpus is extracted using 10-fold cross-validation.
Three values of the learning rate (0.1, 0.01 and 0.001) have been tried on this development corpus and
the best one has been kept.
The number of epochs, up to a maximum of 500 epochs, is also tuned on this development corpus for each experiment.
Furthermore, because some badly initialized weights may fail to converge,
every experiment on the development corpus is run twice with different random initializations, and
the best one on the development set is kept.
Running only twice does not guarantee that at least one training is correct, but it reduces the chances of
failure, without increasing too much the computation requirements.
All other hyper-parameters have been set a priori to reasonable values: 100-dim LSTM hidden state size, 100-dim word embeddings, 0.4 dropout, ReLU activations.

\section{Related work}

Pluwak~\shortcite{pluwak} demonstrates that some expressions of sentiments might not be detected with traditional methods
of opinion mining, and that exploiting dialog acts may partly solve this challenge.
Nevertheless, very few works have investigated this joint modeling of dialog acts and opinion mining:
Clavel and Callejas~\shortcite{clavel} study the impact and usage of both dialog act recognition and sentiment analysis in
human-agent conversational platforms and affective conversational interfaces,
while Novielli and Strapparava~\shortcite{novielli} perform a dialog act clustering of a lexicon and study the emotional load of dialog acts.
The authors further try to improve dialog act recognition by exploiting affective lexicon, but without convincing results.
Conversely, Herzig et al.~\shortcite{herzig} include dialog features (e.g., time elapsed between turns...) into an emotion classifier,
but do not model the dialog and emotion jointly.
Boyer et al.~\shortcite{boyer} exploit affects that are visually expressed on faces to better classify dialog acts.

While many works have investigated sentiment analysis on Twitter~\cite{semeval},
a few works only handle dialog act recognition on Twitter, such as in
\cite{Vosoughi}, \cite{ritter}, \cite{ForsythandMartell2007} and \cite{scheffler}.
Each of these works have annotated their own corpus, because, to the best of our knowledge, no large corpus with dialog act annotations on Twitter exist.
Zarisheva and Scheffler~\shortcite{scheffler} describe a detailed and precise procedure for annotating and evaluating dialog acts on German tweets,
which we have tried to reproduce as closely as possible for English.
Hence, we have manually segmented our own corpus into functional segments -~following the ISO definition of a functional segment \cite{iso}~-
and assumed that this gold segmentation is known during training and testing of our models.
We have also linearized the dialog trees on Mastodon by splitting and reforming every branch of a dialog tree from the root of the tree
as a complete independent dialog.
However, because several such dialogues may share some posts at the beginning of dialogues, we have paid attention that our train,
dev and test splits never contain the same sub-dialog, and are thus completely independent.
We finally took inspiration from the conclusions of~\cite{scheffler}, where it is shown that dialog act recognition is
quite reliable for small taxonomies to design our own taxonomy adapted from the new ISO-standard~\cite{iso}.

A joint model of sentiments and dialog acts is proposed in~\cite{kim18}.
It exploits one convolutional network per task, which output vectors are concatenated and passed to one classifier per task.
It makes strong simplifying assumptions to merge the tasks, in particular:
\begin{itemize}
\item The dialog act at time $t$ does not depend on sentiments, but does depend on the dialog act at time $t-1$;
\item The sentiment at time $t$ only depends on the sentiment and dialog act at time $t$ (no Markov hypothesis);
\end{itemize}
Despite these strong hypothesis, the authors report better results when considering jointly dialog acts and sentiments on
a Korean corpus. As far as we know, their corpus and code is not distributed, which makes comparative evaluations
with this model difficult.
Compared to this work, our proposed network jointly models both tasks at deeper layers, just
after the word embeddings layer. Furthermore, our hypothesis are weaker, because both labels at time $t$ depend on
both tasks at time $t$ and $t-1$, and recurrence is applied at both word and dialog levels.
Finally, we distribute our corpus and software under an open-source and free license to make future comparison easier.

\section{Experimental validation}

Model evaluation is performed with metrics used in related works:
\begin{itemize}
\item For sentiment recognition, following SemEval 2016~\cite{semeval16}, this is the macro-average of the positive and negative F1 scores;
\item For dialog act recognition, following~\cite{scheffler}, this is the average of the dialog-act specific F1 scores weighted by the prevalence of each dialog act.
\end{itemize}

The model has been written in pytorch.
The source code, along with all the data used in this work, is released as open-source and is available at {\url{https://github.com/cerisara/DialogSentimentMastodon}}.

\subsection{Multi-task experiments}
\label{sec:mt}

\begin{figure}[h!]
\begin{center}
\includegraphics[width=8cm]{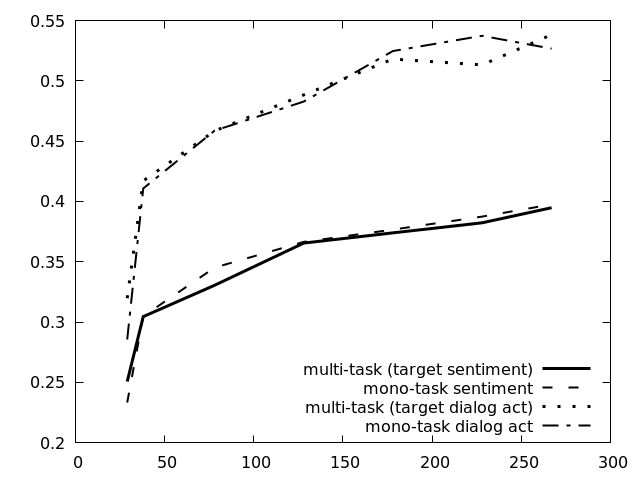}
\end{center}
	\caption[]{Sentiment and dialog act F1s as a function of the number of dialogues used for training.}
	\label{fig:xpmt}
\end{figure}

The development corpus is composed of 28 dialogues (average size across all 10 folds).
We experiment next with a training corpus of varying size: 1, 10, 50, 100, 150, 200 and the full set of 239 dialogues.
The x-axis in Figures~\ref{fig:xpmt} and~\ref{fig:xpre} are labeled with the total number of annotated dialogues
used at training time (train + dev set).
The number of epochs for training is tuned on the development set by optimizing either the sentiment loss ({\bf target sentiment})
or the dialog act loss ({\bf target dialog act}).

In Figure~\ref{fig:xpmt}, both the mono-task and multi-task models have similar performance.
So considering an additional label for the second task does not help, as compared to when only the label of the target
task is given.
This might be due to the fact that the model already captures all available information with a single task,
and the auxiliary label does not bring enough new information to help recognition of the target label.

\subsection{Transfer between tasks}
\label{sec:transfer}

\begin{figure}[h!]
\begin{center}
\includegraphics[width=8cm]{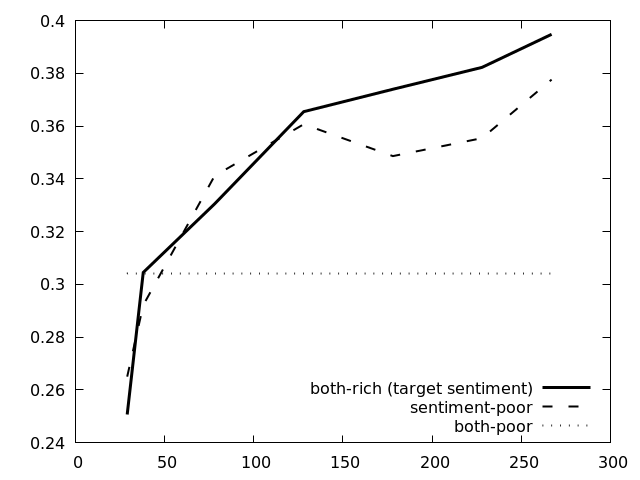}
\end{center}
	\caption[]{Sentiment F1 as a function of the number of dialogues used for training. {\bf both-rich:} both tasks have the same training size; {\bf sentiment-poor:} only 38 dialogues maximum are annotated with sentiment labels; {\bf both-poor:} both tasks are limited to 38 training dialogues.}
	\label{fig:xpre}
\end{figure}

We study next unbalanced cases, when the number of annotations differ between tasks.
Figure~\ref{fig:xpre} compares the evolution of the sentiment analysis F1 score when:
\begin{itemize}
	\item Both tasks are trained on the same number of annotated dialogues ({\bf both-rich}); this is the same experiment as realized in Section~\ref{sec:mt}.
    \item The sentiment recognition task is limited to a maximum of 38 (10 dialogues for training plus the development corpus) annotated dialogues, while
		the training size of the dialog act task is not limited ({\bf sentiment-poor});
	\item Both tasks are limited to only 38 annotated dialogues ({\bf both-poor}).
\end{itemize}

The curves show that
information is largely transferred from the richer dialog act recognition task to
the target sentiment classification task.
Hence, although the {\bf sentiment-poor} and {\bf both-poor} systems 
have only access to 38 dialogues annotated with sentiment labels,
the accuracy of the {\bf sentiment-poor} model keeps on increasing when additional dialog act labels
are considered.
On the right, when the full dialog act training corpus is used, its accuracy is quite close to
the one obtained with all sentiment labels, 
and is better than the {\bf both-poor} model by a large margin, thanks to multi-task transfer learning.

\begin{figure}[h!]
\begin{center}
\includegraphics[width=8cm]{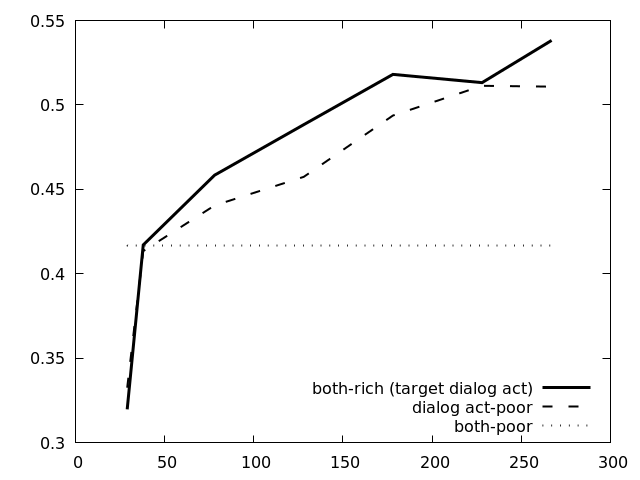}
\end{center}
	\caption[]{Dialog act F1 as a function of the number of dialogues used for training. {\bf both-rich:} both tasks have the same training size; {\bf dialog act-poor:} only 38 dialogues maximum are annotated with dialog act labels; {\bf both-poor:} both tasks are limited to 38 training dialogues.}
	\label{fig:xpreda}
\end{figure}

Conversely, we also validate transfer learning from a rich sentiment recognition task to a poor dialog act recognition task.
The resulting curves are shown in Figure~\ref{fig:xpreda}.

Similar gains in performance are obtained through transfer learning for the dialog act recognition task than
for the sentiment analysis task.
We can thus conclude that transfer learning is efficient between both tasks in both directions.

\section{Analysis}
\label{sec:analysis}

Beyond quantitative experiments that demonstrate transfer learning between dialog act and
sentiment recognition, we analyze next some properties of our Mastodon corpus.

\subsection{Dynamics of sentiments and dialog acts}

In the course of a dialog, the sentiment globally changes at a slower rate than the dialog act.
While dialog acts change nearly at every segment, 
often, a single sentiment is expressed per dialog, sometimes two or three, but rarely more.
This is understandable, but this also implies that the correlation between both tasks is not very strong,
otherwise, they would have a much similar dynamic.
On the other hand, they are not completely independent, as shown in the transfer learning experiments.
So we try and exhibit next some patterns where correlation between tasks is explicit in the corpus.

\subsection{Both tasks are sparsely correlated}
\begin{figure}[h!]
\begin{center}
\includegraphics[width=8cm]{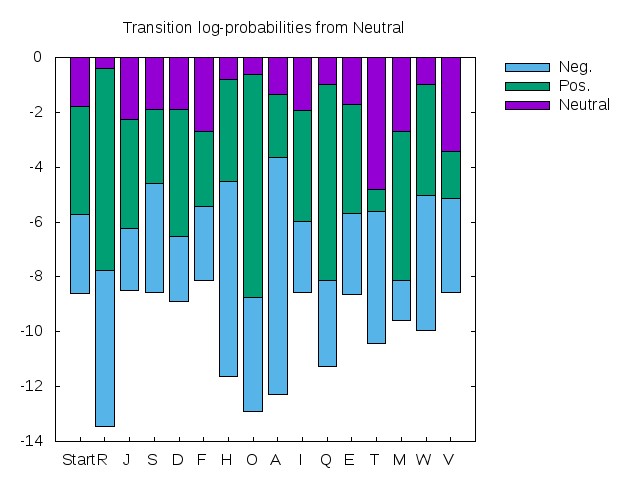}
\caption[]{$\log p(s_t|d_t,s_{t-1}=neutral)$. The dialog acts $d_t$ are listed at the bottom of each histogram column. Because log-probabilities are plotted, boxes with the smallest height are the most likely. The start of a dialog, which has no previous sentiment, is a special case that is added on the left.}
\label{fig:e}
\end{center}
\end{figure}

\begin{figure}[h!]
\begin{center}
\includegraphics[width=7cm]{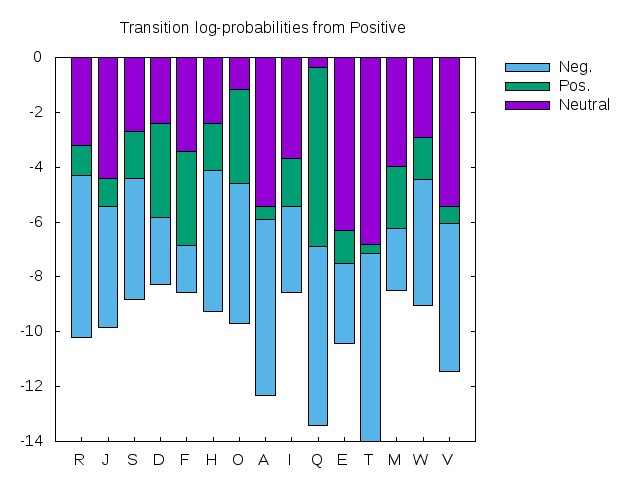}
\includegraphics[width=7cm]{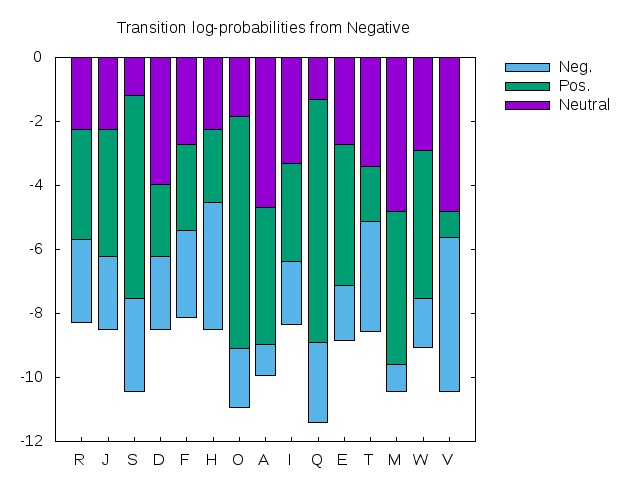}
\caption[]{$\log p(s_t|d_t,s_{t-1}=positive)$ (left) and $\log p(s_t|d_t,s_{t-1}=negative)$ (right) }
\label{fig:pn}
\end{center}
\end{figure}

In order to validate our initial hypothesis that, in the course of a dialog, sentiments may be correlated
to agreements and disagreements, we have computed the transition log-probabilities
between the previous sentiment $s_{t-1}$ and the current sentiment $s_{t}$ as a function of the current dialog act $d_t$.
These log-probabilities are shown respectively for $s_{t-1}=$ neutral, positive and negative in Figures~\ref{fig:e} and \ref{fig:pn}.

We observe that:
\begin{itemize}
    \item With agreements ({\bf A}), a positive sentiment mainly stays positive and a negative sentiment mainly stays negative.
    \item With disagreements ({\bf D}), a positive sentiment becomes either neutral or negative, a negative sentiment becomes either positive or stays negative, a neutral sentiment either stays neutral or becomes negative.
\end{itemize}
These observations globally follow our intuition, even though more complex patterns than expected seem to occur.

Another observation concerns the evolution of sentiments from the start to the end of a dialog.
Previous studies have shown that, with regard to dialogues, the sentiment on social media is related to:
\begin{itemize}
    \item The topic of discussion: for instance, discussions about politics tend to have more negative sentiments~\cite{sentitopic};
    \item The length of the dialog: negative sentiments tend to lead to longer dialogues~\cite{sentilong}, which we suggest may be also related to
        the correlation between sentiments and disagreements that we have observed in this work; indeed, an increased proportion of disagreements
        may lead to longer dialogues.
\end{itemize}
Based on our corpus, another pattern seems to emerge, which is in favor of negative sentiments at the beginning of dialogues and
positive sentiments at the end.
This is intuitively plausible, as at least a small proportion of the sources of discords are likely to be resolved during dialogues, but this
interpretation is still to be confirmed on a more diverse set of corpora.

\section{Conclusion}

We have studied a multi-task model for joint sentiment 
and dialog act recognition on social media.
We have shown that transfer learning is quite efficient when the number
of annotated labels for one task is smaller than for the other task.
We have further analyzed the correlation between both tasks and shown
that although there is enough mutual information to enable transfer learning,
both tasks are not strongly correlated globally.
They are actually characterized by different dynamics, but we have nevertheless
found a few specific patterns that exhibit a strong correlation.
All these studies are realized on a new corpus extracted from the Mastodon
social network.
Conversely to other corpora based on Twitter, this corpus enables reproducible experiments and
both the annotated corpus and source code of our model are available with an open-source license on github.
Obviously, the privacy issues concerned with extracting corpora from social media are not only related to
licenses, and other aspects must be considered, including dissemination, usage and ethical considerations with
respect to citizen privacy.
Even though the proposed approach does not totally solve all such issues, it initiates an original alternative to the current dominant
experimental methodology that hopefully leads to a better experimental reproducibility for natural language researches.

\section*{Acknowledgments}

The authors thank the ``Programme d'Investissements d'Avenir'' of the French government, the French National Research Agency (ANR) and the Lorraine Universit\'e d'Excellence (LUE) initiative for funding.
Experiments presented in this paper were carried out using the Grid'5000 testbed, supported by a scientific interest group hosted by Inria and including CNRS, RENATER and several Universities as well as other organizations (see {\url{https://www.grid5000.fr}}).

% HAL collaboration field: GRID5000
% HAL collaboration field: LUE-UL 

\bibliographystyle{acl}
\bibliography{dasent.bib}

\end{document}